\documentclass{article}

\usepackage[nonatbib, final]{neurips_2021}
\usepackage[numbers,sort&compress]{natbib}

\usepackage[utf8]{inputenc} 
\usepackage[T1]{fontenc}    
\usepackage{hyperref}       
\usepackage{url}            
\usepackage{booktabs}       
\usepackage{amsfonts}       
\usepackage{nicefrac}       
\usepackage{microtype}      
\usepackage{amsthm}
\usepackage{algorithm}
\usepackage{algpseudocode}
\usepackage{mathtools}
\usepackage{graphicx}
\usepackage{subcaption}
\usepackage{tabularx}
\usepackage{dsfont}
\usepackage{multirow}
\usepackage{wrapfig}

\usepackage{color}

\usepackage{multicol}

\usepackage{caption}

\usepackage{enumitem}

\usepackage[dvipsnames]{xcolor}

 \usepackage{tikz}
    \usetikzlibrary{trees, shapes,arrows,automata,positioning,cd}
    \tikzset{
      cfgedge/.style   = {black, ->, >=stealth},
      forward/.style = { blue, ->, >=angle 45},
      backward/.style = { red, densely dashed, ->, >=latex' },
      backwardleft/.style = { red, densely dashed, <-, >=latex' },
    }
    
\DeclareMathOperator*{\argmin}{\arg\!\min}

\DeclareMathOperator*{\argmax}{\arg\!\max}

\DeclarePairedDelimiter{\ceil}{\lceil}{\rceil}

\theoremstyle{definition}
\newtheorem{definition}{Definition}[section]

\newtheorem{theorem}{Theorem}[section]

\title{Fairness Degrading Adversarial Attacks Against Clustering Algorithms}

\author{%
  Anshuman Chhabra\textsuperscript{*}, Adish Singla\textsuperscript{\textdagger}, Prasant Mohapatra\textsuperscript{*}\\
  \textsuperscript{*} University of California, Davis, \texttt{\{chhabra, pmohapatra\}@ucdavis.edu}\\
   \textsuperscript{\textdagger} MPI-SWS, \texttt{adishs@mpi-sws.org}\\
}

\begin{document}

\captionsetup[table]{font=footnotesize, skip=0pt}
\captionsetup[figure]{font=footnotesize, skip=0pt}

\maketitle

\begin{abstract}
Clustering algorithms are ubiquitous in modern data science pipelines, and are utilized in numerous fields ranging from biology to facility location. Due to their widespread use, especially in societal resource allocation problems, recent research has aimed at making clustering algorithms fair, with great success. Furthermore, it has also been shown that clustering algorithms, much like other machine learning algorithms, are susceptible to adversarial attacks where a malicious entity seeks to subvert the performance of the learning algorithm. However, despite these known vulnerabilities, there has been no research undertaken that investigates \textit{fairness degrading adversarial attacks} for clustering. We seek to bridge this gap by formulating a generalized attack optimization problem aimed at worsening the group-level fairness of centroid-based clustering algorithms. As a first step, we propose a fairness degrading attack algorithm for k-median clustering that operates under a \textit{whitebox} threat model-- where the clustering algorithm, fairness notion, and the input dataset are known to the adversary. We provide empirical results as well as theoretical analysis for our simple attack algorithm, and find that the addition of the generated adversarial samples can lead to significantly lower fairness values. In this manner, we aim to motivate fairness degrading adversarial attacks as a direction for future research in fair clustering.

\end{abstract}

\section{Introduction}

Clustering algorithms are unsupervised machine learning algorithms that are utilized in numerous application domains, generally to group similar but unlabeled data together. Some of these application fields include biology \cite{seo2002interactively}, facility location \cite{jung2019center}, computer vision \cite{frigui1999robust}, among others. Clustering algorithms essentially find patterns from the given dataset, and then using those to generate output clusters. Therefore, clustering algorithms, much like other machine learning algorithms, can amplify bias against certain minority protected groups and discriminate against them, if the dataset itself contains biases against certain groups \cite{access}.

This concern has recently given rise to the notion of \textit{group-level} fairness in clustering, with numerous works extending the original ideas first proposed by Chierichetti et al \cite{chierichetti2017fair}. These aim to either \textit{eliminate} biases from the data prior to clustering (pre-processing) \cite{chierichetti2017fair, schmidt2018fair}, change the clustering algorithm itself to make it \textit{fair} (in-processing) \cite{ziko2019clustering, kleindessner2019guarantees, chen2019proportionally}, or \textit{modify} the output clusters to make them fair (post-processing) \cite{bera2019fair, davidsonmaking}. We refer the reader to \cite{access} for a more detailed overview on the different approaches used for fair clustering. In supervised learning and other machine learning fields, along with a number of different fairness intervention approaches proposed, there has been work undertaken to also study adversarial attacks that can exacerbate algorithmic bias and unfairness \cite{nanda2021fairness, mehrabi2020exacerbating, solans2020poisoning, chang2020adversarial}. In stark contrast, while there have been many approaches for fair clustering proposed, very few works have studied adversarial attacks on clustering, and none have proposed adversarial attacks that degrade fairness. 

Recent work has shown \cite{chhabra2020suspicion, cina2021black} that clustering algorithms can indeed \textit{miscluster} (lead to worsening performance) via the addition of adversarial input samples. However, most of this work has been localized to adversarial attacks that seek to change the output clusters in a non-deterministic fashion-- the goal of the adversary is to increase the number of misclustered labels compared to the original clusters prior to the attack. In this work, we take a first step towards proposing adversarial attacks that instead degrade clustering fairness, and make clustering algorithms more unfair. We propose a simple algorithm for the widely utilized k-median centroid-based clustering algorithm, under a \textit{whitebox} full knowledge threat model. To the best of our knowledge, this is the first work that studies adversarial attacks against clustering with the aim of reducing fairness, thus tackling one of the open problems first proposed in \cite{access}. To summarize, we make the following contributions:
\begin{itemize}[wide]
    \item We propose the first fairness degrading adversarial attack against clustering-- specifically for k-median clustering where the adversary operates under a \textit{whitebox} threat model (Sections 3-4).
    \item We provide theoretical as well as empirical analysis of our simple algorithm and demonstrate that it can effectively reduce clustering fairness on a number of real-world datasets (Section 5).
    \item We provide insights and possible directions for future work related to fairness degrading adversarial attacks against clustering algorithms (Section 6).

\end{itemize}

\section{Related Works}

\textbf{Adversarial Attacks Against Clustering.} While research on adversarial attacks against clustering algorithms is generally sparse, recent work has shown that clustering algorithms are vulnerable to blackbox adversarial attacks that increase the misclustering rate of the model \cite{chhabra2020suspicion, cina2021black}. A blackbox attack is essentially a worst-case scenario for the attacker. Thus, the attack algorithms proposed by \cite{chhabra2020suspicion, cina2021black, biggio, biggio_malware, crussell2015attacking} demonstrate that the working of clustering algorithms can indeed be adversely impacted through poisoned data, which motivates our work. Note that since most of the aforementioned work specifically considers lowering performance as the adversary's objective, the attack algorithms cannot be directly utilized to degrade clustering fairness. As we will show in subsequent sections, the attack optimization for degrading clustering fairness leads to an NP-Hard bi-level problem, which has not been considered in clustering literature before.

\textbf{Fair Clustering.} Researchers have recently studied clustering algorithms with fairness guarantees, or proposed fair variants to existing clustering algorithms. Most of this work has looked at centroid-based clustering, and fair algorithmic variants for the k-center and k-median clustering objectives were first proposed by Chierichetti et al \citep{chierichetti2017fair}. Many works have followed that improve upon these ideas, either in terms of approximation rates \citep{backurs2019scalable, bercea2018cost}, allowing for multiple protected groups \citep{bercea2018cost, rosner2018privacy}, considering other clustering objectives \citep{kleindessner2019guarantees, schmidt2018fair}, or a combination of these, among others. Some works also look at alternate notions of fairness for clustering algorithms \citep{chen2019proportionally, ghadiri2020fair}.

\looseness-1 \textbf{Fairness Degrading Adversarial Attacks in Machine Learning.} Fairness degrading attacks have been proposed for supervised learning and deep learning models \cite{nanda2021fairness}. Similar to our work in this paper, most fairness degrading attack problems for these learning models are also bi-level optimization problems \cite{mehrabi2020exacerbating, solans2020poisoning}, which are generally challenging to solve. Moreover, due to the non-trivial nature of the problem, most of this work generally considers more empirical analysis and lacks theoretical guarantees. In our work, we seek to provide strong theoretical justifications for why our attack algorithm is highly effective in reducing clustering fairness, along with experimental results.

\section{Attack Description and Threat Model}

\subsection{Problem Statement}

We now present a formulation for the fairness degrading adversarial attack against clustering algorithms. Let $A_i$ and $A^i$ denote the $i$-th row and $i$-th column of a matrix $A$. We make the following assumptions regarding the class of clustering algorithms we study in this paper-- 1) we only consider centroid-based clustering algorithms (such as k-center \cite{hochbaum1985best}); 2) we only consider \textit{hard} cluster assignments, where data samples can only belong to one cluster and not have partial memberships across many clusters; 3) we restrict our study to clustering algorithms which take the number of clusters as input (such as k-means \cite{lloyd}, k-median \cite{kaufmanl1987clustering}, etc.) and do not determine them as part of the clustering process (DBSCAN \cite{dbscan}, hierarchical clustering \cite{dasgupta2016cost}, etc.). 

We denote the original dataset as $X \in \mathbb{R}^{n \times d}$. For generality, we write the optimization problem in terms of an arbitrary clustering model $\mathcal{M}$, which takes in as input a dataset (such as $X$), and the number of clusters (denoted by $k \leq n$), and outputs a set of $k$ centroids denoted by $\mu \in \mathbb{R}^{k \times d}$. Data samples can then be assigned a cluster label $i$ using the centroid $\mu_i \in \mu$ which is closest to them. 

\looseness-1 For the fairness notions, 1) we consider only \textit{group-level} fairness notions, and 2) notions that are specific to centroid-based clustering (as we are studying only centroid-based clustering algorithms). Many other clustering algorithms and fairness notions exist, but we defer their study to future work. Also, let the number of protected groups (such as gender, race, etc.) be $f \in \mathbb{N}$ and $\mathcal{G}(X, i)$ be a function that outputs a set containing all the points in $X$ that belong to the protected group $i \in [f]$. Note that data samples can only belong to one group at a time. We can thus generate a superset that contains all points in $X$ that belong to all protected groups, denoted by $G = \{G_i\}_{i=1}^{f}$ where $G_i = \mathcal{G}(X, i)$. To make these aforementioned assumptions explicit analytically, we denote the fairness notion using a fairness utility function $\mathcal{U}_{F}$, which operates on the centroid set $\mu$ and the protected group set $G$, and outputs a scalar utility value. Higher utility values denote more fair models.

The attacker aims to add a certain number of adversarial samples, denoted by $X'$ to the original dataset, such that the fairness utility function $\mathcal{U}_F$ is minimized. That is, the adversary aims to shift the original centroids $\mu$ to $\mu^{\textrm{adv}}$ by having the clustering model run on the combined dataset $X \cup X'$, such that $\mathcal{U}_F(\mu^{\textrm{adv}}, G)$ is minimized. Note here that the fairness utility function is still evaluated on the original protected group memberships $G$ since adversarial samples cannot have protected group memberships (an algorithmically generated data sample cannot be of a particular ethnicity, for example). However, even if adversarial samples could have protected group membership labels, the adversarial attack problem then becomes trivial-- it is now a label-spoofing problem, and is no longer related to clustering or machine learning. Finally, since the attacker has limited resources, there is a cost associated with the adversarial attack, denoted by $\mathcal{C}_{A}$. Here, $\mathcal{C}_{A}$ operates on the set of adversarial samples $X'$ and outputs a scalar value. This cost must be less than some threshold value $\epsilon$ where $\epsilon \geq 0$. We can now write the attack problem as an optimization problem for a fixed $X, G, k, \epsilon$:

\begin{equation*}
 \begin{aligned}
\min_{X', \, \mu^{\textrm{adv}}} \quad & \mathcal{U}_F(\mu^{\textrm{adv}}, G) \\
\textrm{s.t.} \quad & \mu^{\textrm{adv}} = \mathcal{M}(X \cup X', k), \\
& \mathcal{C}_{A}(X') \leq \epsilon\\
\end{aligned}   
\end{equation*}

We have now defined the attack problem we are considering in the paper, and we motivate all the aforementioned design choices using a threat model next. Also, as the above formulation is general, we can utilize a variety of definitions for centroid-based clustering using $\mathcal{M}$, as well as fairness notions using $\mathcal{U}_F$. A number of different functions can also be used for the attacker cost $\mathcal{C}_A$. Interestingly, the above optimization problem is a bi-level optimization \cite{colson2007overview}, similar to attack optimizations formulated for fairness degrading attacks for supervised learning models \cite{solans2020poisoning, mehrabi2020exacerbating}.

\subsection{Threat Model}

We now describe the threat model under which the adversary functions, among other details. Since we are considering clustering models, certain aspects of the threat model are predecided. For example, in the paper we consider training-time attacks, where the learning model is modified as a result of the adversarial data addition. Training-time attacks are an intuitive choice for clustering algorithms, as the training and testing sets are the same for clustering algorithms (due to the absence of class labels).  
A testing-time attack could be considered and would constitute using a nearest-centroid search to assign testing set samples to clusters. However, as this is different from attacking just the clustering approach, we defer this to future work.

The threat model is assumed to be as follows:
\begin{enumerate}
    \item As a first approach, we consider a \textit{whitebox} case scenario where the adversary knows the clustering objective being used (and associated parameters), as well as the dataset. Moreover, the adversary also knows the fairness notion that is being employed. Finally, the adversary can query and obtain current cluster centroids at any time. From the defender's perspective this is the worst case, and thus, we seek to demonstrate that for a given clustering algorithm, there can be highly effective fairness degrading attacks that can be launched by an adversary.
    \item As mentioned before, the attacker carries out a training-time attack, in that the clustering model $\mathcal{M}$ is modified/optimized during the adversarial optimization process.
    \item The goal of the attacker is to minimize the fairness utility function $\mathcal{U}_F$ on the original dataset while minimizing their attack cost $\mathcal{C}_A$. 
\end{enumerate}

\subsubsection{Motivating the Threat Model} 

This can be motivated by considering a scenario where a production-level system (such as Microsoft's TAY Chatbot \cite{caton2020fairness}) is using a clustering algorithm in the back-end, and new data samples are provided by users in an online fashion which the model learns from (TAY utilized users' tweets as part of the learning process). To ensure that their model is not becoming unfair, the system designers use a small subset of the data with known group-level memberships to measure the group-level fairness of the model based on some fairness metric.\footnote{Interestingly, Microsoft did not have such a safeguard built in, which resulted in the chatbot becoming racist and misogynistic.}

Given that an adversary exists that has access to this original data subset (in our formulation this is $X$), the learning model $\mathcal{M}$, as well as the fairness utility function $\mathcal{U}_F$, the adversary can introduce data points $X'$ to degrade the fairness guarantees. If the number of samples introduced by the adversary are minimal yet the degradation in fairness is significant, such an attack can be hard to detect and rectify.

\subsubsection{Function Definitions}

We now discuss the choices for the clustering model ($\mathcal{M}$), the fairness utility function ($\mathcal{U}_F$), and the attacker's cost ($\mathcal{C}_A$) considered in the paper:

\textbf{The Clustering Model.} For the centroid-based clustering model $\mathcal{M}$ we consider the k-median clustering objective \cite{kaufmanl1987clustering}. Let the distance metric being used for clustering be denoted as $d : X \times X \rightarrow \mathbb{R}^{\geq 0}$. The goal of k-median clustering then is to partition the input dataset into $k$ clusters, while minimizing intra-cluster dissimilarity and maximizing inter-cluster similarity. For this, the algorithm chooses $k$ centroids for each cluster in a manner such that the distance metric between the cluster centroid and points in its partition is minimized. For k-median, the centers are also \textit{exemplars}, and are chosen from the set of points that make up the original dataset. Define $\Omega(\mu, X) = \sum_{x\in X}\min_{\mu_i \in \mu}d(x, \mu_i)$. The k-median clustering objective is then defined using $\Omega$:
\begin{equation*}
    \mathcal{M}_{\textrm{k-median}}(X,k) = \argmin_{\mu' \in \mathbb{R}^{k \times d}} \Omega(\mu', X) \textrm{ s.t. } \mu' \subseteq X
\end{equation*}

\textbf{The Fairness Notion.} Since as part of our whitebox threat model, we assume explicit analytical knowledge of the fairness notion and fairness utility function $\mathcal{U}_F$, our proposed attack algorithm should work with any fairness notion appropriately defined for k-median clustering. In the rest of the paper and our experiments, we consider the generalized \textit{balance} fairness notion \cite{bera2019fair}. Balance is the most commonly used group-level fairness notion for clustering \cite{access}, and is suitably defined for k-median clustering \cite{access}. Let $\rho_{a,b}$ denote the ratio between proportion of data points that belong to protected group $b \in [f]$, present in the overall dataset $X$ and present just in the cluster $a \in [k]$. This ratio $\rho_{a,b}$ can be calculated using the protected group set $G$ and knowledge of clusters (using centroid set $\mu$). Then, we can define the balance utility fairness function $\mathcal{U}_F(\mu, G) \in [0,1]$ as follows:

\begin{equation*}
    \mathcal{U}_F(\mu, G) = \min_{a \in [k], b \in [f]} \min \{ \frac{1}{\rho_{a,b}}, \rho_{a,b} \}
\end{equation*}

As can be seen in the definition above, higher balance values signify more fair clustering models (the highest possible value being 1 and the lowest being 0). Thus, higher utility signifies more fairness.

\textbf{The Adversary's Cost.} The attacker wishes to add as few points as possible, as we assume that adding more points gives the defender more knowledge of the attacker's subversion efforts. This choice of attack cost also becomes more clear with the working of our attack algorithm which we cover in later sections. Thus, we define the attack cost in a simple manner as $\mathcal{C}_A(X') = |X'|$. We defer the study of more nuanced attack costs as part of future work.

\textbf{The Specific Attack Optimization Problem.} Combining all the above definitions, we can rewrite the original attack optimization problem:

\begin{equation*}
 \begin{aligned}
\min_{X', \, \mu^{\textrm{adv}}} \quad & \mathcal{U}_F(\mu^{\textrm{adv}}, G) \\
\textrm{s.t.} \quad\, & \mu^{\textrm{adv}} = \argmin_{\mu' \in \mathbb{R}^{k \times d}, \mu' \subseteq X} \Omega(\mu', X \cup X'), \\
& |X'| \leq \epsilon\\
\end{aligned}   
\end{equation*}

\section{The Proposed Fairness Degrading Attack}

We now describe our proposed approach for solving the bi-level attack optimization described in the previous section. Since we are attempting to solve a bi-level problem, we employ a two-step process that can provide us with feasible solutions. Let the original centers on $X$ be denoted as $\mu^0$. As a preprocessing first step, we find possible \textit{adversarial} cluster centers $\mu^{\textrm{adv}}$ that decrease the fairness utility, ie $\mathcal{U}_F(\mu^0) \geq \mathcal{U}_F(\mu^{\textrm{adv}})$. Then, as part of our main attack algorithm, we aim to add adversarial data samples to the original dataset so as to \textit{realize} these centers in the actual clustering. The preprocessing step is shown as Subroutine 1 (\textsc{FindAdversarialCentroids}). The main attack algorithm is shown as Algorithm 1 and uses the adversarial centers obtained via Subroutine 1. 

\begin{algorithm}[ht!]
\caption*{\textbf{Subroutine 1: }\textsc{FindAdversarialCentroids}}
\textbf{\underline{Input}:} $X, k, G, \mathcal{U}_F$\\
 \textbf{\underline{Output}:} $\mu^{\textrm{adv}}$
 \begin{algorithmic}[1]
 \State \textbf{select} $k$ centroids as set $\mu'$ s.t. $\mathcal{U}_F(\mu', G)$ is minimal
 \While{fairness utility $\mathcal{U}_F(\mu', G)$ decreases}
 \For{each pair $(m, p) \in \mu' \times (X \setminus \mu')$}
 \State \textbf{consider} the swap between $p$ and $m$:
 \State \quad\, \textbf{compute} change in $\mathcal{U}_F(\mu', G)$
 \State \textbf{if} change in $\mathcal{U}_F(\mu', G)$ is the current best:
 \State \quad\, \textbf{set} $(m^*,p^*) \leftarrow (m,p)$ 
 \EndFor
 \State \textbf{if} best swap decreases $\mathcal{U}_F(\mu',G)$ \textbf{swap} $p^*$ and $m^*$ 
 \EndWhile
 \State \textbf{return} $\mu'$ as $\mu^{\textrm{adv}}$
 \end{algorithmic}
\end{algorithm}

\begin{algorithm}[ht!]
\caption{Proposed Fairness Degrading Attack Algorithm}
 \textbf{\underline{Input}:} $X, k, G, \epsilon, \Omega, \mathcal{U}_F$\\
 \textbf{\underline{Output}:} $X'$
 \begin{algorithmic}[1]
 \State $\mu^{\textrm{adv}} \leftarrow $ \textsc{FindAdversarialCentroids}($X, k, G, \mathcal{U}_F$)
 \State \textbf{find} $\hat{\mu}$ s.t. $\max_{x' \in X}d(x, x') \leq d(x, \hat{\mu}), \forall x \in X \setminus \hat{\mu}$
 \State \textbf{define} $\Theta(\mu, X) = \Omega(\{ \hat{\mu}\}, X) - \Omega(\{ \hat{\mu}\} \cup \mu, X)$ for a center set $\mu$ of any length, and any $X$
 \State \textbf{initialize} $X' \leftarrow \emptyset$
 \While{$true$}
 \State \textbf{if} $|X'| > \epsilon$ \textbf{then return} $\emptyset$
 \State \textbf{set} $\mu^*_0 \leftarrow \emptyset, \mu' \leftarrow \emptyset$
 \State \textbf{obtain} $\mu^*_i = \mu^*_{i-1} \cup \{\argmax_{x \in X\cup X'} \{\Theta(\mu^*\cup x, X\cup X') - \Theta(\mu^*, X \cup X')\}\}$, $\forall i\in [k]$
 \State \textbf{if} $\mu^*_i = \mu^{\textrm{adv}}_i$, $\forall i\in [k]$ \textbf{then return $X'$}
 \For{each cluster $i \in [k]$}
 \While{$\argmax_{x \in X\cup X'} \{\Theta(\mu'\cup x , X\cup X') - \Theta(\mu', X \cup X')\} \neq \mu^{\textrm{adv}}_i$}
 \State $X' \leftarrow X' \cup \{ \mu_i^{\textrm{adv}}\}$
 \EndWhile
 \State $\mu' \leftarrow \mu' \cup \{ \mu^{\textrm{adv}}_i\}$
\EndFor
\EndWhile
\State \textbf{return} $X'$
 \end{algorithmic}
\end{algorithm}

\textbf{Subroutine 1.} The proposed approach is based on the Partitioning-Around-Medoids (PAM) algorithm \cite{kaufmanl1987clustering} which is widely used as a heuristic for solving the NP-Hard k-median objective. Initially we select the set $\mu'$ by greedily and iteratively picking points from $X$ that decrease the fairness utility $\mathcal{U}_F(\mu',G)$ the most (line 1). After obtaining this temporary center set, we begin the while loop which iterates till the fairness utility does not change (line 2). In this loop, we find the best medoid, non-medoid pair $(m^*,p^*)$ to swap, so that this leads to the largest decrease in fairness utility (lines 3-10). Now, since \textsc{FindAdversarialCentroids} is based on PAM, it also inherits the latter's convergence and stability properties. To summarize, if $\mu^{\textrm{adv}}$ is returned via \textsc{FindAdversarialCentroids}, and $\mu^{0}$ are the set of original centers obtained by just running PAM (k-median algorithm) on $X$, then we know that $\mathcal{U}_F(\mu^{\textrm{adv}}) \leq \mathcal{U}_F(\mu^0)$ since swaps are made in a greedy fashion.

\looseness-1 \textbf{Algorithm 1.} We first provide a high-level overview of the working of Algorithm 1, and then a more detailed explanation for our approach. Basically, to minimize the k-median clustering objective, we transform the minimization objective in terms of $\Omega$ to a submodular maximization problem. This allows us to solve the k-median problem in a structured manner while ensuring the centroid set eventually becomes the adversarial center set $\mu^{\textrm{adv}}$ obtained from Subroutine 1. Next, since we can easily obtain the center set for the k-median problem by maximizing this submodular function, we iteratively check the centroids obtained and keep adding points at adversarial center locations till the adversarial centroids are the obtained k-median centroids/medoids. If we add more points than the maximum allowable attacker cost ($\epsilon$) we return with no solution for $X'$. In this case, $\epsilon$ would have to be increased to a larger value that will allow us to compute a feasible $X'$. Finally, an intuitive justification for why adding adversarial samples to $X'$ directly at centroids locations of $\mu^{\textrm{adv}}$ helps us obtain $\mu^{\textrm{adv}}$ as the optimal center set, can be derived from the k-median clustering objective itself. Essentially, by adding points at centers in $\mu^{\textrm{adv}}$, clustering costs will increase for all other candidate center sets except for centers in $\mu^{\textrm{adv}}$ (refer to the defintion for $\mathcal{M}_{\textrm{k-median}}$). Moreover, there has to be an upper bound on the number of adversarial samples required to be added till the centroid set $\mu^{\textrm{adv}}$ is realized. This claim can be understood by the following argument-- considering we added an infinite number of points at each given adversarial center, these would be the optimal centers as there is infinite mass at these locations, making them the candidate center (medoid/exemplar) set with the lowest clustering cost for solving the k-median problem. 

Thus, the goal of the attack algorithm is to ensure that the optimal clustering center set obtained on $X\cup X'$ is very close to the adversarial centers obtained by Subroutine 1 (line 1). We also assume, without loss of generality, that adversarial center sets are ordered and can be accessed using the cluster label. Then, as mentioned above-- even though the original k-medoids clustering objective using $\Omega$ is hard to optimize for, it can be converted into a submodular function to ease minimization. We do this by first defining a \textit{phantom} center $\hat{\mu}$ that satisfies the condition in line 2. Then using $\hat{\mu}$ we can convert the k-median objective-- in line 3, we define the modified clustering objective function $\Theta$. This is clearly a monotone submodular function, and it can be trivially verified that maximizing $\Theta$ is equivalent to minimizing $\Omega$, thus minimizing the k-median clustering objective $\mathcal{M}_{\textrm{k-median}}$. We can thus exploit the submodularity of $\Theta$ to generate the adversarial sample set $X'$ close to the optimal clustering solution. We initialize $X'$ to an empty set in line 4. Then we begin a while loop (lines 4-14) that returns the empty set if we exhaust the attacker's resources (line 6). Subsequent to this, we check to see how close the adversarial centers are to being a good approximation to the k-medoids objective for the current $X'$, so as to eventually return a feasible solution. 

\looseness-1 For this we utilize the seminal greedy approach of Nemhauser \cite{nemhauser1978analysis} for submodular function maximization (utilized in line 8). If $\mu^*_0$ is initialized as an empty set, this can be defined using the iterative step: $\mu^*_i = \mu^*_{i-1} \cup \{\argmax_{x \in X\cup X'} \{\Theta(\mu^*\cup x, X\cup X') - \Theta(\mu^*, X \cup X')\}\}$, $\forall i\in [k]$. Therefore, the center set obtained via this iterative formulation is close to an optimal solution for the k-median clustering problem. Thus we check to see if the adversarial centers are the same as obtained via this greedy approach and if so, return the adversarial center set $X'$ (line 9). The remainder of the algorithm (lines 10-15) basically keeps adding adversarial samples at locations $\mu^{\textrm{adv}}_{i}$ for $i\in[k]$ to $X'$, till the centers obtained by greedy submodular maximization are equal to $\mu^{\textrm{adv}}$. Here the for loop (line 10) aims to do this sequentially for each cluster center, and the while loop (line 11) adds points to $X'$ till the greedy submodular maximization condition is met. However, note every time we move on to a new cluster, adversarial data addition might invalidate the greedy selection for a previous cluster. That is, an adversarial center obtained as a cluster center in the previous iteration of the for loop (line 10) might no longer be the cluster center in the current iteration due to new adversarial points added at the next adversarial center location. This is why we have the initial while loop (line 5), as it allows us to repeat this process till the overall checking condition is met (line 9). To summarize, we are attempting to find the adversarial sample set $X'$ such that when we utilize the greedy approach of \cite{nemhauser1978analysis} for maximizing $\Theta$ on $X\cup X'$ we obtain the adversarial center set $\mu^{\textrm{adv}}$. 

Also, as mentioned before, we can also discuss the convergence of the while loop (line 5) in Algorithm 1, ie, can we even obtain $\mu^{\textrm{adv}}$ as the cluster centers by adding points at these adversarial center locations? We can answer this question in the affirmative with a simple argument: consider adding a large number of points at each adversarial center for all $k$ centers by adding points at these locations in $X'$. Since possible options for centers are finite (they have to be a subset of $X\cup X'$), a large enough point mass at each adversarial center will eventually lead to that point being selected as the medoid/center. This can be verified by observing the definition for $\mathcal{M}_{\textrm{k-median}}$ and $\Omega$ as well. Thus, Algorithm 1 converges.

We can now also give theoretical justification for the performance of Algorithm 1. We present this result as Theorem \ref{approx}. Essentially, this results tells us that the adversarial samples $X'$ obtained using Algorithm 1 lead to the adversarial center set $\mu^{\textrm{adv}}$ being a $1-1/e$ approximation for the optimal k-median clustering on the combined dataset $X\cup X'$. That is, Algorithm 1 generates adversarial samples that allow for the adversarial centers (which are known to decrease fairness) to be close to the optimal k-median clustering solution, thus leading to a successful fairness degrading attack.

\begin{theorem} \label{approx}
\textit{Let $\mu^{\textrm{OPT}}$ be the optimal centers for the k-median clustering problem on $X\cup X'$ (defined using $\Theta$) where the adversarial sample set $X'$ is obtained via Algorithm 1 and $\mu^{\textrm{adv}}$ denotes the adversarial center set obtained via Subroutine 1, then $\mu^{\textrm{adv}}$ is a $(1-1/e)$ approximation to $\mu^{\textrm{OPT}}$, ie, $\Theta(\mu^{\textrm{adv}}, X\cup X') \geq (1 - \frac{1}{e})\Theta(\mu^{\textrm{OPT}}, X\cup X')$.}
\end{theorem}

Under a commonly made assumption regarding the clusters being \textit{well-separated} \cite{tan2016introduction}, we can even provide an upper bound for the maximum attacker cost, given by $\epsilon$. This is shown in Theorem \ref{bound}. However, we first define well-separated clusters \cite{tan2016introduction}:

\begin{definition}(\textbf{\textit{Well-separated Clusters}}). \textit{These are defined for a given $k$ and dataset $X$ as a set of cluster partitions $\{P_1, P_2,..., P_k\}$ on the dataset, s.t. points belonging to each $P_i \subset X$ are closer to each other than any points belonging to $P_j \subset X$ where $i \neq j, \forall i,j \in [k]$.}
\end{definition}

\begin{theorem}\label{bound}
\textit{For a well-separated k-median clustering problem and $\sigma_i = \min_{x' \in P_i}\sum_{x \in P_i}d(x,x')$, the maximum size of the adversarial sample set $X'$ obtained using Algorithm 1 is bounded as:}
\begin{equation*}
\begin{aligned}
    \epsilon \leq \sum_{i \in [k]} |P_i| \ceil[\bigg]{\frac{(\sum_{x \in P_i}d(x, \mu_i^{\textrm{adv}}) - \sigma_i )}{\min_{y \in P_i\setminus \mu_i^{\textrm{adv}}}d(y, \mu_i^{\textrm{adv}})}}\\
\end{aligned}
\end{equation*}
\end{theorem}

\section{Results}
In this section, we provide empirical results showcasing the working of Algorithm 1 on real datasets. We consider the following 3 UCI datasets for experiments:

1) \textbf{Adult} \citep{kohavi1996scaling}. The UCI \textit{adult} dataset consists of 10000 samples with 5 features (\textit{age, education-num, final-weight, capital-gain, hours-per-week}) chosen for clustering. The original dataset contains 48842 samples, but like previous works \citep{chierichetti2017fair}, \citep{abraham2019fairness} we remove missing values and undersample to 10000 samples. The protected groups are \textit{white, black, asian-pac-islander, amer-indian-eskimo, other} signifying the feature \textit{race}. Therefore, $n \times d = 10000 \times 5$.\\
2) \textbf{Bank} \citep{moro2014data}. The UCI \texttt{bank} dataset contains 45211 samples and 3 features (\textit{age, balance, duration}), i.e. $n \times d = 45211 \times 3$. The protected groups for \texttt{bank} are \textit{married, single, divorced} signifying \textit{marital status}.\\
3) \textbf{Creditcard} \citep{yeh2009comparisons}. The UCI \texttt{creditcard} data comprises of 30000 samples ($n$) with 23 features ($d$). The protected groups are \textit{higher} and \textit{lower} education signifying the feature \textit{education}.

\begin{table}[ht!]
\centering
\fontsize{8}{8}\selectfont
\caption{Results for Algorithm 1 on real datasets.}
\begin{tabular}{cccccc}
\hline
\#Clusters & Dataset & Pre-attack $\mathcal{U}_F$ & Post-attack $\mathcal{U}_F$ & \% Decrease in $\mathcal{U}_F$ & $\mathcal{C}_A(X') = |X'|$ \\ \hline
\multirow{3}{*}{$k = 2$} & Bank & 0.3037 & 0.0 & 100 & 0.0022$|X|$  \\ 
 & Creditcard & 0.8612 & 0.0 & 100 & 0.0067$|X|$  \\ 
 & Adult & 0.5668 & 0.0 & 100 & 0.0600$|X|$ \\ 
 \hline
\multirow{3}{*}{$k = 3$} & Bank & 0.2978 & 0.0 & 100 & 0.0288$|X|$  \\ 
 & Creditcard & 0.7580 & 0.0 & 100 & 0.0100$|X|$  \\ 
 & Adult & 0.0 & - & - & - \\ 
 \hline
\multirow{3}{*}{$k = 4$} & Bank & 0.2829 & 0.0 & 100 & 0.0420$|X|$ \\ 
 & Creditcard & 0.4640 & 0.0 & 100 & 0.0300$|X|$  \\ 
 & Adult & 0.0 & - & - & - \\ 
 \hline
\end{tabular}
\end{table}

In the experiments that follow, we vary the number of clusters $k$ from 2 to 4 and then run Algorithm 1 to carry out the fairness degrading attack on k-median clustering for each value of $k$. Furthermore, as mentioned before we utilize balance as the fairness utility function. Also, we set $\epsilon$ to a large number for these experiments to showcase how much adversarial sample addition would actually be required to realize the adversarial centers found by Subroutine 1. In Table 1, we present the results obtained. We find that we can effectively decrease fairness under our threat model, while incurring minimal attacker cost, that is, very few adversarial samples are added. As can be seen in Table 1, for each choice of $k$ we obtain the fairness utility function $\mathcal{U}_F$ before the attack (centroids obtained by clustering on $X$) and after the attack (centroids obtained by clustering on $X \cup X'$). In all cases we find that we can minimize balance to make it the lowest possible value (0.0) with only very few adversarial samples added. This can be seen in the column showcasing attacker cost $\mathcal{C}_A(X')$ where the size of the adversarial sample set $X'$ is a small fraction of the original dataset $X$. Also note that for $k=3$ and $k=4$ for the Adult dataset, the original clustering itself leads to a fairness value of 0.0, which cannot be minimized further, and thus there is no reason to attack.


\section{Discussion}

\textbf{Extending and improving attack algorithms.} Currently, Algorithm 1 is very simplistic as we are just adding adversarial points at fixed locations. While these locations are not known to the defender, the adversarial data could still be found and removed if the data addition is excessive (although in our experiments we find that data addition is very minimal). However, better algorithms for both Subroutine 1 and Algorithm 1 can be proposed for future work. Moreover, an algorithm that directly tackles the bi-level attack optimization might be even more effective, albeit non-trivial to devise. Furthermore, stronger theoretical guarantees could be provided in future work-- the approximation ratio currently provided as part of Algorithm 1 is based on the modified clustering objective function $\Theta$ and not the original function $\Omega$. An extension of our work could be to propose an attack algorithm for which theoretical guarantees can be given based on the original clustering objective.

\textbf{Attack transferability to fair clustering approaches.} An important research question relates to whether our attack (and associated adversarial samples) can be thwarted through fair clustering approaches, forming a \textit{defense} approach against fairness degrading attacks. In the context of adversarial sample transferability \cite{papernot2016transferability}, this can be rephrased as-- \textit{do adversarial samples generated by a fairness degrading adversarial attack algorithm transfer across fair clustering approaches and worsen their performance?} This question is not trivial to answer, as all fair clustering approaches require protected group memberships for samples in the data, and in our current threat model adversarial samples cannot possess protected group labels. For future work, this problem could be considered further, possibly by modifying the threat model.

\section{Conclusion}

In this paper, we propose the first fairness degrading adversarial attack against clustering algorithms. We consider the popular k-median clustering algorithm and the balance fairness metric, and provide an attack algorithm that can easily degrade fairness while incurring minimal attacker cost (formulated as the number of adversarial samples added). We provide theoretical results for the performance of our attack algorithm under an assumed whitebox threat model, which is the worst-case scenario for the defender. We also undertake experiments on real data, and find that our attack is highly effective, and can easily reduce fairness while adding very few adversarial samples.

\bibliographystyle{unsrt}



\end{document}


\captionsetup[table]{font=footnotesize, skip=0pt}
\captionsetup[figure]{font=footnotesize, skip=0pt}

\maketitle

\section{Proofs}

\subsection{Proof for Theorem 4.1}

\begin{manualtheorem}{4.1}\label{approx}
\textit{Let $\mu^{\textrm{OPT}}$ be the optimal centers for the k-median clustering problem on $X\cup X'$ (defined using $\Theta$) where the adversarial sample set $X'$ is obtained via Algorithm 1 and $\mu^{\textrm{adv}}$ denotes the adversarial center set obtained via Subroutine 1, then $\mu^{\textrm{adv}}$ is a $(1-1/e)$ approximation to $\mu^{\textrm{OPT}}$, ie, $\Theta(\mu^{\textrm{adv}}, X\cup X') \geq (1 - \frac{1}{e})\Theta(\mu^{\textrm{OPT}}, X\cup X')$.}
\end{manualtheorem}

\begin{proof}
The proof for this theoretical guarantee of Algorithm 1 rests upon the seminal work of Nemhauser on submodular function maximization \cite{nemhauser1978analysis}. Nemhauser proposed a greedy selection strategy for maximizing a submodular function defined over a set. For completeness, we first describe this strategy in detail and then demonstrate how it plays a role in Algorithm 1. Assume a given submodular function $R(Q)$ which operates on the set $Q$. For this problem, we wish to select $k$ elements from $S$ to populate set $Q$ such that $R(Q)$ is maximized. Also for our selection strategy, let $Q^i$ denote the $i$-th iteration consisting of selecting an element from $S$ for $Q$. Moreover, let $Q^0$ be the empty set $\emptyset$. Then, Nemhauser proposed the following iterative strategy:
\begin{equation*}
    Q^i = Q^{i-1} \cup \{\argmax_{s \in S} R(Q^{i-1} \cup s) - R(Q^{i-1})\}
\end{equation*}

If a $k$ sized set $Q^k$ is picked using this strategy, and $Q^{\textrm{OPT}}$ is the optimal $k$ sized solution to the above problem, Nehmauser proved the following result: $R(Q^k) \geq (1-1/e) R(Q^{\textrm{OPT}})$.

We can now use the above result to prove the theorem statement. For Algorithm 1, in lines 2-3 we essentially transform the k-median minimization problem to a submodular maximization problem. To do this, in line 2 we first define a phantom exemplar $\hat{\mu}$ and use it to define a new clustering objective $\Theta$ in line 3. As maximizing $\Theta$ is equivalent to minimizing $\Omega$ and $\Theta$ is a submodular function, we will aim to maximize it using Nemhauser's greedy selection approach. Essentially we are using the greedy approach in line 8 with $\Theta$ as the submodular function and $X\cup X'$ as the set we are picking elements from: $\mu^*_i = \mu^*_{i-1} \cup \{\argmax_{x \in X\cup X'} \{\Theta(\mu^*\cup x, X\cup X') - \Theta(\mu^*, X \cup X')\}\}$, $\forall i\in [k]$. 

Then Algorithm 1 iterates, and only returns a feasible set if the set $\mu^*$ obtained using the greedy approach (line 8) is the same as the adversarial centroid set $\mu^{\textrm{adv}}$ obtained using Subroutine 1. Therefore, we can now use Nemhauser's result to verify the theorem statement and complete the proof simply by replacing the following: $Q^k \rightarrow \mu^{\textrm{adv}}, R \rightarrow \Theta, Q^{\textrm{OPT}} \rightarrow \mu^{\textrm{OPT}}, S \rightarrow X\cup X'$.

\end{proof}

\subsection{Proof for Theorem 4.2}

\begin{manualtheorem}{4.2}\label{bound}
\textit{For a well-separated k-median clustering problem and $\sigma_i = \min_{x' \in P_i}\sum_{x \in P_i}d(x,x')$, the maximum size of the adversarial sample set $X'$ obtained using Algorithm 1 is bounded as:}
\begin{equation*}
\begin{aligned}
    \epsilon \leq \sum_{i \in [k]} |P_i| \ceil[\bigg]{\frac{(\sum_{x \in P_i}d(x, \mu_i^{\textrm{adv}}) - \sigma_i )}{\min_{y \in P_i\setminus \mu_i^{\textrm{adv}}}d(y, \mu_i^{\textrm{adv}})}}\\
\end{aligned}
\end{equation*}
\end{manualtheorem}

\begin{proof}

By well-defined clusters in the theorem statement, we are assuming that the clustering problem is well-defined, with each of the $k$ clusters being far apart from each other, making the clustering output static and deterministic. Now, to prove the upper-bound on $|X'|$, that is, $\epsilon$. Algorithm 1 basically adds points to all target cluster centers till they are chosen as the medoid/centroid of that cluster. Essentially, for each cluster partition, this means that we increase whatever the current cluster center cost is (in terms of the clustering objective), by adding points in $X'$ such that the clustering cost with the target center is smaller than that. Consider a cluster $i$ with cluster partition $P_i$. Let the original cluster center be $\mu_i = \argmin_{x' \in P_i}\sum_{x \in P_i}d(x,x')$, $\sigma_i = \min_{x' \in P_i}\sum_{x \in P_i}d(x,x') = \sum_{x \in P_i}d(x,\mu_i)$, and $\sigma_i^{\textrm{adv}} = \sum_{x \in P_i}d(x,\mu_i^{\textrm{adv}})$. 

To make $\mu_i$ ineligible as the center, we would need to increase the clustering cost with $\mu_i$ by adding points in $X'$ at $\mu_i^{\textrm{adv}}$, till the clustering cost with $\mu_i^{\textrm{adv}}$ as the center is lesser than it. Each time we add a point to $X'$ at $\mu_i^{\textrm{adv}}$ we increase the clustering cost with center $\mu_i$ by $d(\mu_i^{\textrm{adv}}, \mu_i)$. The number of points $l_i$ to be added to cluster $i$ to make $\mu_i$ ineligible as center is then:
\begin{equation*}
    l_i = \ceil[\bigg]{\frac{\sigma_i^{\textrm{adv}} - \sigma_i}{d(\mu_i^{\textrm{adv}} - \mu_i)}}
\end{equation*}

However, just because $l_i$ points have been added does not mean that $\mu_i^{\textrm{adv}}$ will be chosen as the next medoid for cluster $i$. While $\mu_i$ will definitely be ineligible, some new cluster center $\mu_i'$ might exist in $P_i$ that has lower cost than $\mu_i^{\textrm{adv}}$. Therefore, to effectively have $\mu_i^{\textrm{adv}}$ be chosen as the cluster center, we would have to add another $l_i'$ points to make $\mu_i'$ ineligible as center, and would have to do the same for any other intermediate centers till $\mu_i^{\textrm{adv}}$ is chosen as center. Similar to before, let $\sigma_i' = \sum_{x \in P_i}d(x,\mu_i')$. 

Now, we also know that $\sigma_i < \sigma_i'$ for any intermediate center $\mu_i'$ chosen after $\mu_i$. This is simple to see-- the clustering cost for the point $\mu_i'$ when no points were added to $X'$ and $\mu_i$ was the center was greater than $\sigma_i$ by the definition of $\mu_i$. Then, as new points were added to $X'$ this cost could only grow, and become $\sigma_i'$. Since, $\sigma_i < \sigma_i'$, it holds that:
\begin{equation}
\sigma_i^{\textrm{adv}} - \sigma_i > \sigma_i^{\textrm{adv}} - \sigma_i'.     
\end{equation}

On the other hand, as the intermediate centers are clearly moving towards $\mu_i^{\textrm{adv}}$ we have for any intermediate center $\mu_i'$ chosen after $\mu_i$:
\begin{equation}
d(\mu_i^{\textrm{adv}}, \mu_i) > d(\mu_i^{\textrm{adv}}, \mu_i')
\end{equation}

 Thus, using the above inequality (2) we can see that the difference in centers will be largest for the last intermediate center before $\mu^{\textrm{adv}}_i$ becomes the center. This difference in centers can be denoted as $\min_{y \in P_i\setminus \mu_i^{\textrm{adv}}}d(y, \mu_i^{\textrm{adv}})$. 
 
 Using inequalities (1) and (2), we can write down an upper-bound for $l_i'$ which is the number of points to be added to make any intermediate center $\mu_i'$ chosen after $\mu_i$ ineligible:

\begin{equation}
    l_i' \leq \ceil[\bigg]{\frac{\sigma_i^{\textrm{adv}} - \sigma_i}{\min_{y \in P_i\setminus \mu_i^{\textrm{adv}}}d(y, \mu_i^{\textrm{adv}})}}
\end{equation}

To ascertain the number of points to be added to the cluster $i$ we would need to sum $l_i'$ over all such intermediate centers $\mu_i'$. However, the exact number of intermediate centers is not easily known, as this is largely data-dependent. Therefore, since the clusters are well-separated and each cluster $i$ contains $|P_i|$ points, the maximum possible candidates for centers are capped at $|P_i|$. Therefore, the total number of points to be added to cluster $i$ for the target center to be chosen as medoid is $|X'|$:

\begin{equation}
    |X'| \leq |P_i| l_i'
\end{equation}

Since clusters are well-separated, we can sum over all such $|X'|$ given in equation (4), $\forall i \in [k]$, and expand $l_i'$ using equation (3) and also write $\sigma_i^{\textrm{adv}} = \sum_{x \in P_i}||x-\mu_i^{\textrm{adv}}||^2$ to obtain the maximum size of $X'$ and hence, $\epsilon$:

\begin{equation*}
    \epsilon \leq \sum_{i \in [k]} |P_i| \ceil[\bigg]{\frac{(\sum_{x \in P_i}d(x,\mu_i^{\textrm{adv}}) - \sigma_i )}{\min_{y \in P_i\setminus \mu_i^{\textrm{adv}}}d(y, \mu_i^{\textrm{adv}})}}
\end{equation*}

\end{proof}

\bibliographystyle{unsrtnat.bst}
\bibliography{dblp_ref.bib}